\documentclass[runningheads]{llncs}
\usepackage[T1]{fontenc}
\usepackage{amsmath}
\usepackage{graphicx}
\usepackage{amsfonts}
\usepackage{booktabs}
\usepackage{multirow}
\usepackage{array}
\usepackage{tabularx}
\usepackage{ragged2e}
\usepackage{caption}
\usepackage[misc]{ifsym}
\begin{document}
\title{STGDPM:Vessel Trajectory Prediction with Spatio-Temporal Graph Diffusion Probabilistic Model}
\titlerunning{Vessel Trajectory Prediction with STG Diffusion Probabilistic Model}
\author{
Wenzhe Jin \and
Xudong Zhang \and
Haina Tang \textsuperscript{(\Letter)}}
\authorrunning{W, Jin et al.}
\institute{School of Artificial Intelligence, University of Chinese Academy of Sciences, Beijing, China  \\
\email{\{jinwenzhe23,zhangxudong221\}@mails.ucas.ac.cn, hntang@ucas.ac.cn}
}
\maketitle              % typeset the header of the contribution
\begin{abstract}
Vessel trajectory prediction is a critical component for ensuring maritime traffic safety and avoiding collisions. Due to the inherent uncertainty in vessel behavior, trajectory prediction systems must adopt a multimodal approach to accurately model potential future motion states. However, existing vessel trajectory prediction methods lack the ability to comprehensively model behavioral multi-modality. To better capture multimodal behavior in interactive scenarios, we propose modeling interactions as dynamic graphs, replacing traditional aggregation-based techniques that rely on vessel states. By leveraging the natural multimodal capabilities of diffusion models, we frame the trajectory prediction task as an inverse process of motion uncertainty diffusion, wherein uncertainties across potential navigational areas are progressively eliminated until the desired trajectories is produced. In summary, we pioneer the integration of Spatio-Temporal Graph (STG) with diffusion models in ship trajectory prediction. Extensive experiments on real Automatic Identification System (AIS) data validate the superiority of our approach.

\end{abstract}
\section{Introduction}
With the continuous expansion of the global economy, international trade volumes have surged, driving rapid increases in vessel capacity, size, and speed \cite{kaluza2010complex}. In this context, ship safety and security have become increasingly critical, particularly in mitigating collision risks that can lead to casualties, substantial property and cargo losses, and long-term environmental damage \cite{akyuz2015hybrid}. To effectively avoid collision risks between vessels and enhance navigation safety, accurately predicting vessel trajectories in dynamic environments has become a critical technology.

Predicting vessel trajectories presents several challenges:

•	Interaction: Vessel navigation is influenced by surrounding Vessels. Identifying other vessels within visibility range and adjusting trajectories accordingly is a conventional method to avoid collisions.

•	multi-modality: For the same historical trajectory, there can be multiple safe and reasonable future trajectories that correspond.

Traditional trajectory prediction often relies on aggregation-based techniques \cite{alahi2016social,gupta2018social,liu2023qsd} and Transformer-based models \cite{sadeghian2019sophie,yuan2021agentformer} for interaction modeling. Aggregation methods combine features from neighboring entities through operations like summation or averaging, but they may overlook the varying importance of different neighbors, limiting their ability to capture complex interactions. Transformer models, using self-attention, assign dynamic weights to interacting entities, allowing them to model nuanced dependencies. However, standard positional encodings in Transformers are designed for sequential data, which makes them less effective for irregular time intervals or dynamic spatial relationships. While both methods offer some interaction modeling, they struggle to handle complex, dynamic interactions in irregular spatio-temporal contexts, highlighting the need for more advanced approaches.

Due to the inherent strong modeling capabilities and efficiency of Spatio-Temporal Graph (STG) models \cite{mohamed2020social} , STG has demonstrated powerful multi-node sequence prediction capabilities across various domains, including traffic prediction\cite{yu2017spatio}, weather forecasting\cite{tekin2021spatio}, and power forecasting\cite{simeunovic2021spatio}. In the field of vessel trajectory prediction, STG has gradually become mainstream, achieving significant results in related studies\cite{feng2022stgcnn,mohamed2020social,wu2024gl}. Existing STG methods often adopt the binary Gaussian distribution introduced by Social-LSTM\cite{alahi2016social} to represent the multi-modality of trajectories. However, this distribution can only capture multi-modality near the predicted endpoint. In the field of pedestrian trajectory prediction, Social-GAN addresses the challenge of learning trajectory multi-modality by minimizing the loss for k predicted trajectories, effectively preventing the model from converging to a mean distribution that would otherwise approximate the ground truth through label-based learning\cite{gupta2018social}. However, the manifold of the distribution of future paths is may be discontinued and thus cannot be well covered by GAN or CVAE-based methods\cite{dendorfer2021mg}.

Denoising Diffusion Probabilistic Models (DDPMs) are emerging as a promising approach for spatio-temporal graph modeling. Originally developed for image generation and denoising, DDPMs are gaining attention for their ability to model complex, multimodal distributions and capture uncertainty over time and space. In spatio-temporal graph contexts, DDPMs can handle dynamic interactions, generating diverse possible trajectories in uncertain environments. By modeling the gradual diffusion of noise and reversing the process, DDPMs capture intricate spatio-temporal dependencies, making them well-suited for trajectory prediction tasks, where both spatial relationships (e.g., proximity) and temporal factors (e.g., past states) influence future outcomes. However, existing methods, such as aggregation-based approaches and those relying on spatially fixed sensor data, are limited in modeling dynamic spatio-temporal graphs and are only applicable to static data.

Therefore, in order to solve these problems, we combine Spatio-Temporal Graph methods with Denoising Diffusion Probabilistic Models and propose the design of Traj-UGnet for dynamic spatio-temporal graph modeling. In this framework, future vessel positions are treated as particles: under low uncertainty, they converge into a clear path, while under high uncertainty, they disperse across navigable areas. The diffusion process is simulated by progressively adding noise to the trajectory, mimicking the spread of possible paths. By reversing this process, uncertainty is gradually reduced, converting ambiguous regions into more deterministic trajectories. Unlike traditional stochastic methods that introduce randomness via latent noise variables, this approach explicitly simulates the evolution of uncertainty, capturing the multi-modality of vessel movement. Our main contributions are summarized as follows:
\begin{itemize}
    \item To the best of our knowledge, our research is the first to integrate Spatio-Temporal Graph (STG) models with diffusion models for vessel trajectory prediction, effectively capturing complex spatio-temporal dependencies and inherent uncertainties in vessel trajectory data.

    \item We model vessel trajectories as dynamic spatio-temporal graphs and propose a denoising network, traj-UGnet, within the diffusion model to effectively extract their spatio-temporal features.

    \item We conduct extensive experiments on realistic vessel trajectories under different water areas to prove the effectiveness of STGDPM, which demonstrates that STGDPM is better than existing models. Further experiments prove that STGDPM have the ability to model trajectory behavioral multi-modality.
\end{itemize}

\section{Related work}
Most trajectory prediction models treat it as a sequence prediction problem. Related studies indicate \cite{khosroshahi2016surround,wu2020comprehensive} that the unique architecture of Recurrent Neural Networks (RNN) performs excellently in handling long-term dependency issues, especially models like Gated Recurrent Unit (GRU) and Long Short-Term Memory (LSTM) networks \cite{han2019real,capobianco2021deep}. LSTM, through its complex gating mechanisms, effectively addresses the vanishing gradient problem in conventional RNNs.

In addition to temporal sequence modeling, many studies focus on modeling complex spatial interactions. With the rapid development of deep learning technologies, increasingly efficient network architectures have been designed to simulate social interactions. For instance, Social LSTM \cite{alahi2016social} is one of the earliest deep models dedicated to pedestrian trajectory prediction, introducing a social pooling layer to aggregate interaction information from the environment. In SFM-LSTM \cite{liu2022deep} and QSD-LSTM \cite{liu2023qsd}, the Social Force Model (SFM) and Quaternion Ship Domain (QSD) model are integrated into the original LSTM network, enhancing the capability for vessel trajectory prediction. 

Despite the advances in spatio-temporal modeling made by these RNN models, limitations remain. Firstly, these models typically only model time dependency within single nodes, failing to effectively capture temporal correlations between different nodes. In reality, objects are often tightly connected in space and time, failure to encode such spatial dependencies can significantly reduce prediction accuracy \cite{wu2019graph,yu2017spatio}. Consequently, rather than separately modeling time and space, Spatio-Temporal Graph models (STG) have been proposed to jointly model temporal cues and social interactions \cite{feng2022stgcnn,mohamed2020social,wu2024gl}.

Regarding the multimodal nature of trajectories, considering observed information, agents can make multiple reasonable and socially acceptable future predictions. This characteristic makes multimodal trajectory prediction distinct from other data modalities. Due to limited environmental cues and the inherent randomness of motion, models are unlikely to predict a single future trajectory that accurately matches reality. Thus, numerous studies propose stochastic prediction methods to model multimodal future movements. For example, assuming that the position at each timestep follows a bivariate Gaussian distribution, the model's objective is to maximize the likelihood of the true scenario in the predicted distribution via negative log-likelihood loss. This strategy was initially used in Social LSTM \cite{alahi2016social} for deterministic prediction, but Social-GAN \cite{gupta2018social} discarded it due to its non-differentiable position sampling, creating an RNN-based generative model. Other approaches \cite{dendorfer2021mg,gupta2018social,sadeghian2019sophie} use Generative Adversarial Networks (GAN) for multi-modal modeling, while others \cite{salzmann2020trajectron++,yuan2021agentformer} adopt Conditional Variational Autoencoders (CVAE). MID \cite{gu2022stochastic} was proposed for trajectory prediction with controllable diversity, which following the Denoising Diffusion Probabilistic Model (DDPM) \cite{ho2020denoising,sohl2015deep}, latent vector $y_k$ is sampled from a Gaussian distribution with controlled randomness via a parameterized Markov chain.  

DDPM is a class of deep generative models inspired by non-equilibrium thermo-dynamics, initially proposed by Sohl-Dickstein et al. DDPM has gained significant attention due to its advanced performance in various generative tasks, including image generation and audio generation. DDPM learns to denoise the original public distribution toward a specific data distribution via a parameterized Markov chain. DDPM excels at generating multimodal data distributions, capable of handling complex and uncertain situations, and producing multiple plausible predictions. Many works have verified the feasibility of diffusion model in STG prediction \cite{gu2022stochastic,ma2024dm,wen2023diffstg}. In summary, from the perspectives of spatio-temporal modeling and trajectory multi-modality, we designed a diffusion architecture based on traj-UGnet for vessel trajectory prediction.

\section{Problem Definition}

The input of the prediction system is the $N$ history trajectories in a scene such that $\mathbf{x}^i = \{ p_t^i=(lat_{t}^{i},lon_{t}^{i}) | t = -T_{obs}, -T_{obs} + 1, \cdots, 0\}$, $\forall i \in \{1, 2, \cdots, N\}$, where the $ p_t^i$ is The latitude and longitude coordinates recorded by AIS data of vessel $n$ at timestamp $t$, $T_{obs}$ denotes the length of the observed trajectory, and the current timestamp is $t = 0$. Similarly, the predicted future trajectories can be written as $\mathbf{y}^i = \{ p_t^i=(lat_{t}^{i},lon_{t}^{i}) | t = 1, 2, \cdots, T_{pred}\}$. For clarity, we use x and y without the superscript $n$ for the history and future trajectory in the following subsections.

\begin{figure}
\includegraphics[width=\textwidth]{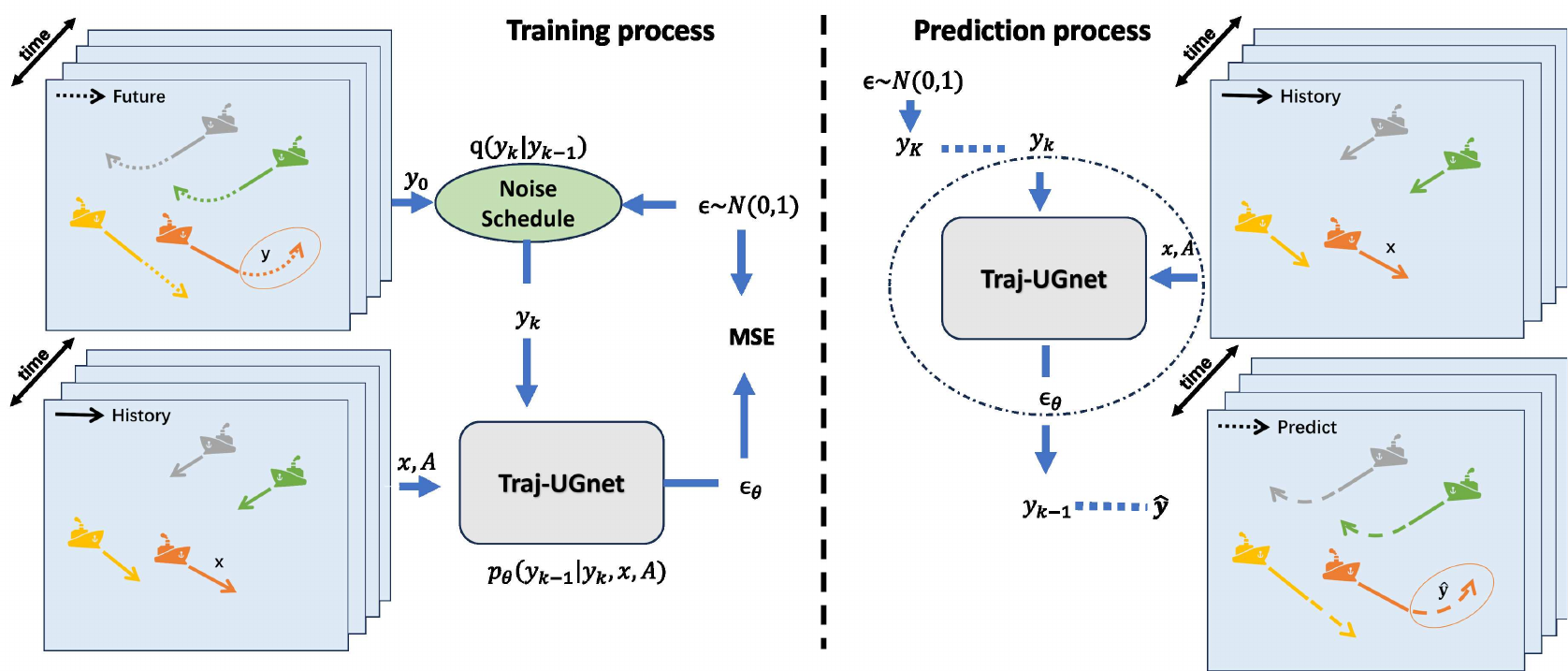}
\caption{STGDMP framework.} 
\label{struction}
\end{figure}

\subsection{Graph Representation of Vessel Trajectories}

We first introduce the construction of the dynamic graph representation of vessel trajectories. We start by constructing a set of spatial graphs $G_t$ representing the locations of vessels in a scene at each time step $t$. $G_t$ is defined as $G_t = (V_t, E_t)$, where $V_t = \left\{p_t^i \mid \forall i \in \{1,\ldots,N\}\right\}$ is the set of vertices of the graph $G_t$, with the number of vertices potentially varying at each time step. $E_t$ is the set of edges within graph $G_t$, which is expressed as $E_t = \left\{e_{t}^{ij} \mid \forall i,j \in \{1,\ldots,N\}\right\}$. $e_{t}^{ij} $ models how strongly two nodes could influence with each other. $e_t^{ij}$s are organized into the weighted adjacency matrix $A_t$. $e_{t}^{ij}$ is defined in equation~\ref{eq:kernel}, with additional details on the $\tau$ for $e_t^{ij}$ discussed in section~\ref{Ablation}. For clarity, we denote the set of dynamic spatial graphs of historical trajectories as $A$ in subsequent sections.

\begin{equation}
    e_{t}^{ij} =
    \begin{cases}
        1/\|p_t^i - p_t^j\|_2 & , 0 < \|p_t^i - p_t^j\|_2 < \tau \\
        0 & , Otherwise.
    \end{cases}
    \label{eq:kernel}
\end{equation}

\section{Method}

In this section, we generalize the popular DDPM to dynamic spatio-temporal graphs and present a novel framework called STGDPM for probabilistic STG forecasting in this section. We describe how to train and predict by this diffusion model shown in fig ~\ref{struction}. Finally, to adapt to dynamic graphs scenes, we propose our denoising network Traj-UGnet.

\subsection{STGDMP framework}

We present the details of the STGDMP framework shown in fig~\ref{struction}. The framework consists of two main stages: training and prediction. In the training process, unlike the original DDPM, which generates images without conditioning, we use a conditional diffusion model to generate future trajectories based on historical information. Specifically, historical trajectories \(x\) and dynamic spatial graphs \(A\) are input into the model, enabling it to leverage past movement patterns and spatial dependencies for more accurate trajectory predictions. Future trajectories \(y_0\) undergo a noise schedule, adding Gaussian noise iteratively to create \(y_k\). The Traj-UGnet model learns to denoise \(y_k\) to estimate the original trajectory by minimizing the mean square error (MSE). In the prediction process, random noise is fed into the model with \(x\) and \(A\), which iteratively generates a predicted trajectory \( \hat{y} \), simulating possible future paths. 

\subsection{Conditional Diffusion Model}

The diffusion and reverse diffusion processes are both formulated as a Markov chain with Gaussian transitions. Specifically, we define the diffusion process as $(y_0,y_1,\cdots,y_K)$, where $K$ represents the maximum number of diffusion steps. This process progressively introduces indeterminacy, transforming the ground truth trajectory into noisy data. Conversely, the reverse process, denoted as $(y_K,y_{K-1},\cdots,y_0)$, incrementally reduces this indeterminacy from $y_K$, ultimately recovering the desired trajectory.

First, we define the posterior distribution for the forward diffusion process, which progresses from \( y_0 \) to \( y_K \), as follows:
\begin{equation}
    q(y_{1:K}|y_0):=\prod_{k=1}^{K} q(y_k|y_{k-1}),
\end{equation}
where each step is represented as:
\begin{equation}
    q(y_k|y_{k-1}):=\mathcal{N}(y_k;\sqrt{1-\beta_k}y_{k-1},\beta_k\textbf{I}),
\end{equation}
with fixed variance schedulers \( \beta_1, \beta_2, \dots, \beta_K \) controlling the scale of noise at each step. Exploiting the properties of Gaussian transitions, let \( \alpha_k = 1 - \beta_k \) and \( \bar\alpha_k = \prod_{s=1}^k \alpha_s \). The distribution of \( y_k \) given \( y_0 \) can then be expressed in closed form as:
\begin{equation}
    q(y_k|y_0):=\mathcal{N}(y_k;\sqrt{\bar\alpha_k}y_0,(1-\bar\alpha_k)\textbf{I}).
\end{equation}
As \( K \) becomes large enough, we approximate \( y_K \sim \mathcal{N}(\textbf{0}, \textbf{I}) \), indicating that, as noise is added, the signal is gradually transformed into a Gaussian noise distribution. This behavior aligns with the principles of non-equilibrium thermodynamics that describe the diffusion process.

Next, we define the generation of trajectories as a reverse diffusion process that starts from the noise distribution. This reverse process is modeled by parameterized Gaussian transitions conditioned on the observed trajectories, formulated as:
\begin{equation}
    \begin{aligned}
    &p_\theta(y_{0:K}|x,A):={p(y_K)}\prod_{k=1}^K {p_\theta(y_{k-1}|y_k,x,A)},\\
    &p_\theta(y_{k-1}|y_k,x,A):=\mathcal{N}(y_{k-1};\mu_\theta(y_k,k,x,A);\Sigma_\theta(y_k,k)),   
    \end{aligned}
\end{equation}
where \( p(y_K) \) is an initial Gaussian noise distribution, and \( \theta \) represents the parameters of the diffusion model. The parameters \( \theta \) are trained using trajectory data, with shared network parameters across all transitions. Following prior work \cite{ho2020denoising}, the variance term in the Gaussian transition can be set as \( \Sigma_\theta(y_k,k) = \sigma_k^2 I = \beta_k I \), which serves as an upper bound on the reverse process entropy and has demonstrated effectiveness in practice \cite{sohl2015deep}. Furthermore, we adopt the parameterization method from \cite{ho2020denoising} to reparameterize:
\begin{equation}
    \mu_\theta(y_k, k, x, A) = \frac{1}{\sqrt{\alpha_k}}(y_k - \frac{\beta_k}{\sqrt{1-\alpha_k}}\epsilon_\theta(y_k, k, x, A)),
\end{equation}
resulting in a simplified loss function:
\begin{equation}
    L(\theta, \varphi) = \mathbb{E}_{\epsilon, y_0, k}\|\epsilon - \epsilon_{(\theta,\varphi)}(y_k, k, x, A)\|,
\end{equation}
where \(\epsilon \sim \mathcal{N}(0, I)\). Given that \( y_0 \) is known during training, and using the relationship \( y_k = \sqrt{\overline{\alpha}_k} y_0 + \sqrt{1 - \overline{\alpha}_k} \epsilon \) from the forward process, the training objective for unconditional generation is specified as:
\begin{equation}
    L(\theta, \varphi) = \mathbb{E}\|\epsilon - \epsilon_{(\theta,\varphi)}(\sqrt{\overline{\alpha}_k}y_0 + \sqrt{1 - \overline{\alpha}_k}\epsilon, k, x, A)\|,
\end{equation}
and training is conducted at each step \( k \in 1, 2, \ldots, K \).

\begin{table}[htbp]
    \centering

    % Algorithm 1: Training of STGDPM
    \resizebox{\columnwidth}{!}{
        \begin{tabular}{p{\dimexpr\columnwidth-4\tabcolsep}}
            \hline
            \multicolumn{1}{l}{\textbf{Algorithm 1.} Training of STGDPM} \\
            \hline
            \hangindent=2em \hangafter=1 \textbf{Input:} Distribution of training data $q(\boldsymbol{y}_0)$, number of diffusion step $K$, variance schedule $\{\beta_1,\cdots,\beta_K\}$, historical vessel trajectories $\boldsymbol{x}$, adjacency matrix $A$. \\
            \textbf{Output:} Trained denoising function $\epsilon_\theta$ \\
            1: \textbf{repeat} \\
            2: \quad $k \sim \text{Uniform}(\{1,\cdots,K\})$, $\boldsymbol{y}_0\sim q(\boldsymbol{y}_0)$ \\
            3: \quad Sample $\boldsymbol{\epsilon}\sim\mathcal{N}(0,I)$ where $\boldsymbol{\epsilon}$'s dimension corresponds to $\boldsymbol{y}_0$ \\
            4: \quad Calculate noisy targets $\boldsymbol{y}_k=\sqrt{\alpha_k}\boldsymbol{y}_0+\sqrt{1-\alpha_k}\boldsymbol{\epsilon}$ \\
            5: \quad Take gradient step $\nabla_{\theta}\|\boldsymbol{\epsilon}-\epsilon_\theta(\boldsymbol{y}_k,k,x,A )\|_2^2$ according to Eq. (8) \\
            6: \textbf{until} converged \\
            \hline
        \end{tabular}
    }

    \vspace{10pt}

    % Algorithm 2: Sampling of STGDPM
    \resizebox{\columnwidth}{!}{
        \begin{tabular}{p{\dimexpr\columnwidth-4\tabcolsep}}
            \hline
            \multicolumn{1}{l}{\textbf{Algorithm 2.} Sampling of STGDPM} \\
            \hline
            \hangindent=2em \hangafter=1 \textbf{Input:}  Historical vessel trajectories $\boldsymbol{x}$, adjacency matrix $A$, trained denoising function $\epsilon_\theta$ \\
            \textbf{Output:} Future forecasting $\boldsymbol{y}$ \\
            1: Sample $\boldsymbol{\epsilon}\sim\mathcal{N}(0,I)$ where $\boldsymbol{\epsilon}$'s dimension corresponds to $\boldsymbol{y}$ \\
            2: \textbf{for } $k=K$ \textbf{to 1 do} \\
            3: \quad Sample $\boldsymbol{y}_{k-1}$ using Eq. (5) by taking $\boldsymbol{x}$ and $A$ as condition \\
            4: \textbf{end for} \\
            5: Return $\boldsymbol{y}$ \\
            \hline
        \end{tabular}
    }
    
\end{table}

\subsection{Inference}

After training the reverse process, we generate plausible trajectories by starting from a Gaussian noise variable \( y_K \sim \mathcal{N}(0, I) \) and applying the reverse process \( p_0 \). Using the reparameterization defined in Equation~\ref{eq:Inference}, we generate trajectories in reverse, from \( y_K \) to \( y_0 \), as follows:
\begin{equation}
     y_{k-1} = \frac{1}{\sqrt{\alpha_k}} \left( y_k - \frac{\beta_k}{\sqrt{1 - \alpha_k}} \epsilon_\theta(y_k, k, x, A) \right) + \sqrt{\beta_k} z,
     \label{eq:Inference}
\end{equation}
where \( z \) is a random variable from a standard Gaussian distribution, and \( \epsilon_\theta \) is the trained network. The inputs to \( \epsilon_\theta \) include the prediction from the previous step \( y_k \), historical vessel trajectories $x$, adjacency matrix $A$, and the current step $k$.

\subsection{Traj-UGnet}
\begin{figure}
\includegraphics[width=\textwidth]{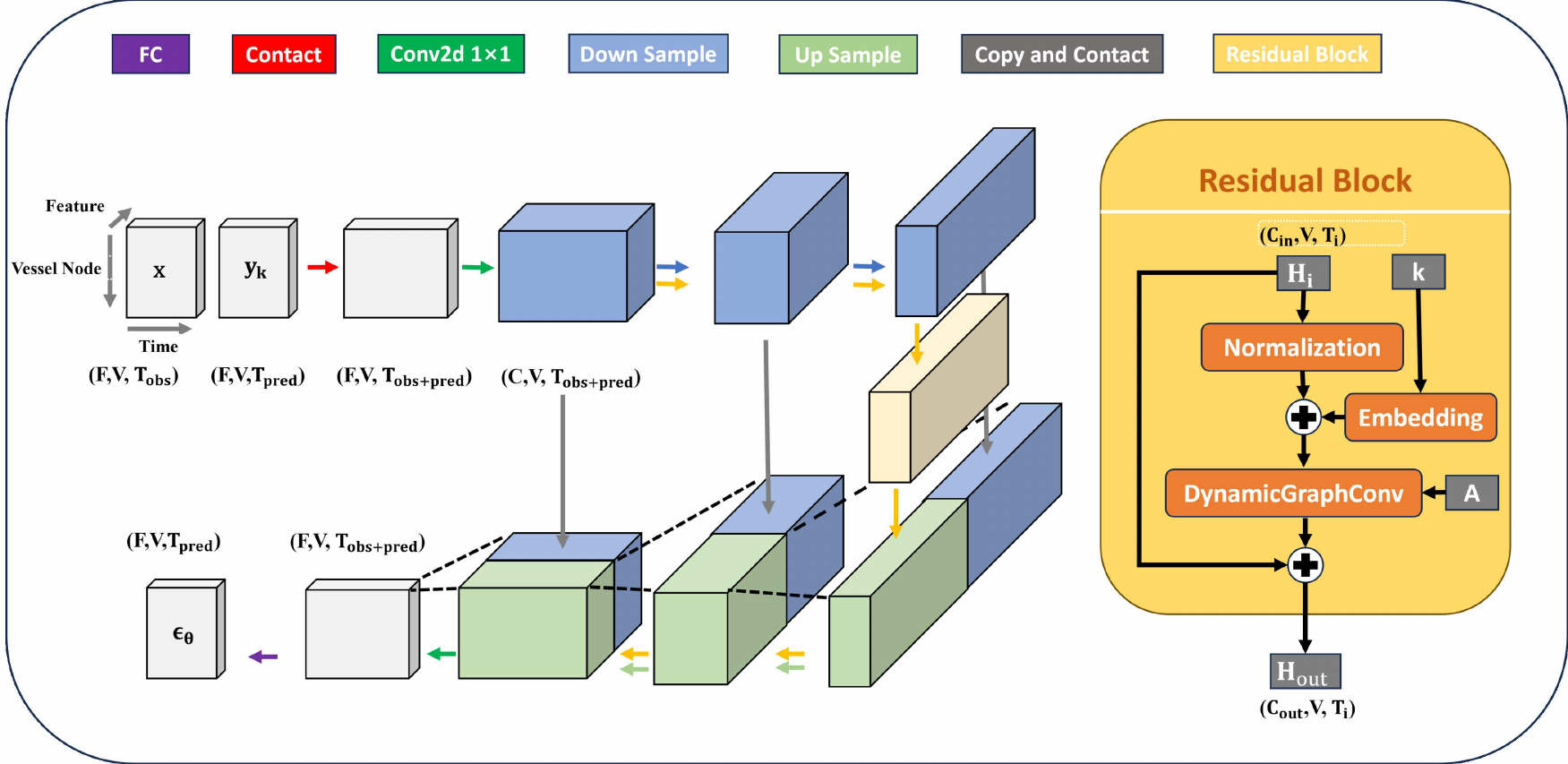}
\caption{Architecture of our Traj-UGnet module. The color of the arrow corresponds to the color of the function module.} \label{trajUGnet}
\end{figure}

The overall process of traj-UGnet is illustrated in the figure~\ref{trajUGnet}. traj-UGnet takes ${y_K}$, conditional information ${x}$, $A$ as inputs, and outputs the noise $\epsilon_\theta$. Traj-UGnet combines ${x}$ and ${y_K}$ and extracts spatio-temporal features through a U-shaped network structure. The core of this U-Net is the residual block, which is used throughout the network. Within each residual block, the DynamicGraphConv module leverages the historical dynamic interaction graph \(A\), further enhancing the model's effectiveness. We will introduce each part in detail in the following.

First, the historical trajectories ${x}\in\mathbb{R}^{F\times V\times T_{obs}}$ are concatenated with $\boldsymbol{y_K}\in\mathbb{R}^{F\times V\times T_{pred}}$ along the time axis become ${h}\in\mathbb{R}^{F\times V\times T_{obs+pred}}$. ${h}$ is then embedded into a deeper-dimensional process variable ${H}\in\mathbb{R}^{C\times V\times T_{pred}}$ through a 1$\times$1 Conv2d kernel, where C is the hidden dimension. 

Next, the model undergoes some down-sampling and up-sampling steps, each down-sampling and up-sampling step is accompanied by multiple residual blocks. each labeled as block $i$. Let $ H_i \in \mathbb{R}^{C_{in} \times V \times T_i} $(where \( H_0 = H \)) denote the input of the i-th Residual Block, where \( T_i \) is the length of time dimension. then $ H_i \in \mathbb{R}^{C_{out} \times V \times T_i}$ is the output. During each downsampling step, $C_{out}$ increases while $T_i$ decreases. In contrast, during each upsampling step, $C_{out}$ decreases and $T_i$ increases. In the intermediate stages between downsampling and upsampling, the dimensionality remains unchanged. At each upsampling step, the input is concatenated with the output from the corresponding downsampling step (see the gray arrow in Figure~\ref{trajUGnet}). 

Finally, the output of the upsampling is then passed through a 1$\times$1 Conv2d kernel, which projects it from $\mathbb{R}^{C_{out} \times V \times T_i}$ back to $\mathbb{R}^{F \times V \times T_{obs+pred}}$. a fully connected (FC) operation is applied to align the time dimension with that of ${y_K}$, producing the final output.

\subsubsection{Residual block.} As shown in Figure~\ref{trajUGnet}, the input ${H}_i$ is first normalized, after which the embedded noise step value $k$ is added (we used the embedding method from \cite{wen2023diffstg}). The resulting value $\hat{H}_i$ is then processed by a DynamicGraphConv to fuse the spatio-temporal features, producing the residual value, which is then added back to the input.
\paragraph{DynamicGraphConv.} We perform spatio-temporal aggregation on graph data represented by a dynamic adjacency matrix \( A \in \mathbb{R}^{T_{obs} \times V \times V} \) and a feature tensor \( \hat{H}_i \in \mathbb{R}^{C_{in} \times V \times T_i} \). First, we normalize the adjacency matrix \( A \) as follows:
\begin{equation}
    \hat{A} = I - D^{-\frac{1}{2}} A D^{-\frac{1}{2}},
\end{equation}
where \( D \) is the degree matrix, and \( I \) is the identity matrix. Next, we aggregate spatio-temporal information using the following operation:
\begin{equation}
   H_c = \hat{A} \hat{H}_i,
\end{equation}
where \( H_c \in \mathbb{R}^{C_{in} \times V \times T_i \times T_{obs}} \). The aggregated feature tensor \( H_c \) is then transformed using learnable convolutional weights \( \theta \in \mathbb{R}^{C_{\text{in}} \times C_{\text{out}} \times T_{\text{obs}}} \) to produce the final feature representation across time and vessels. This transformation is formulated as:
\begin{equation}
   H_{\text{gc}} = \theta H_c + b,
\end{equation}
where \( b \) is a bias term, the output \( H_{\text{gc}} \in \mathbb{R}^{C_{\text{out}} \times V \times T_i} \). Finally, a ReLU activation is applied to the output of the graph convolution layer, with a residual connection that allows the model to retain the original feature information while incorporating the spatio-temporal aggregation.

To summarize, DynamicGraphConv can be expressed by the following formula:
\begin{equation}
H_{\text{out}} = \sigma \left( \theta \hat{A} \hat{H}_i + b + \hat{H}_i \right)
\end{equation}

\begin{figure}
\includegraphics[width=\textwidth]{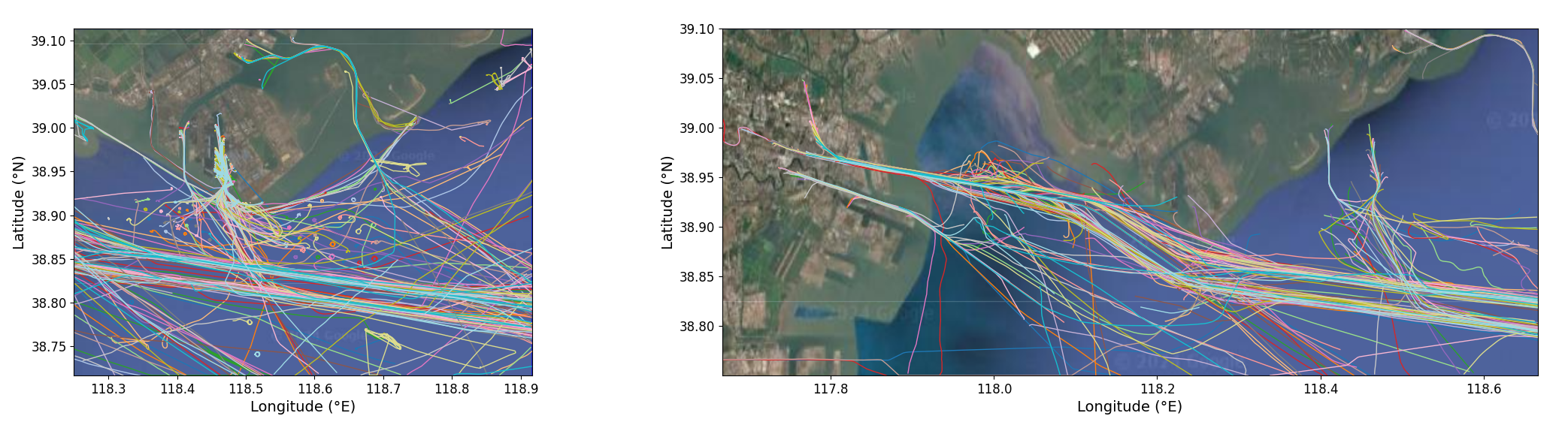}
\caption{From left to right are Caofeidian Waters and Tianjin Port. Tianjin, with its moderate depth and urban location, handles medium-sized container ships, tankers, and ro-ro vessels, making it a key multimodal trade hub for northern China. In contrast, Caofeidian’s deep-water facilities support vessels over 200,000 DWT, focusing on bulk commodities like coal, iron ore, and crude oil, with infrastructure for large-scale handling and transshipment.} \label{fig1}
\end{figure}

\section{Experiment}
\subsection{Dataset}
To evaluate the robustness of the model in varying environments, this study conducts experiments using real vessel trajectory datasets from the Caofeidian Waters (CFW) and Tianjin Port (TJP) \cite{liu2023qsd,liu2022stmgcn} (see Fig.~\ref{fig1}). These datasets include MMSI, timestamp, and geolocation data. Due to limitations inherent in the AIS data trans-mission and reception mechanisms, the raw AIS data contain numerous missing and anomalous values. After preprocessing for data completion and cleansing, the Caofeidian and Tianjin Port datasets comprise 293,636 and 191,106 Trajectories, respectively, along with 5,834 and 7,860 interaction scenarios, each involving two or more vessels(more details see Table.~\ref{table:ais_datasets}).

\begin{table}[htb]
    \centering
    \caption{Summary of AIS Dataset Statistics}
    \label{table:ais_datasets}
    \resizebox{\columnwidth}{!}{
        \begin{tabular}{lcccc}
            \toprule
             \textbf{Region} &  \quad\textbf{Time Period} &  \quad\textbf{Vessel Trajectories} &  \quad\textbf{Interactive Scenes} & \quad \textbf{Max Vessels per Scene} \\
            \midrule
            Caofeidian Waters (CFW) & Jun. 8-10, 2018 & 293,636 & 5,834 & 131 \\
            Tianjin Port (TJP) & Jul. 8-10, 2018 & 191,106 & 7,860 & 64 \\
            \bottomrule
        \end{tabular}
    }
\end{table}

\subsection{Evaluation Metric}
We use two widely adopted evaluation metrics: Average Displacement Error (ADE) and Final Displacement Error (FDE). ADE measures the average error between all ground truth positions and the predicted positions across the trajectory, while FDE measures the error between the final points of the ground truth and predicted trajectories. With a 10-second interval between trajectory points, our model observes 10 steps in the past and forecasts future trajectories at 5, 10, and 15 steps ahead. Given the stochastic nature of our approach, we apply a Best-of-N strategy (N = 20) to compute the final ADE and FDE values. 

% $\hat{p}_t^i$ is the predicted value of the actual trajectory point $p_t^n$ of the corresponding vessel $i$ at time $t$.

% \begin{equation}
%     \mathrm{ADE} = \frac{\sum_{i \in N}\sum_{t \in T_p}|\hat{p}_t^i - p_t^i|2}{N\times T_{pred}}\ 
% \end{equation}

% \begin{equation}
%     \mathrm{FDE} = \frac{\sum_{i \in N}|\hat{p}_t^i - p_t^i|_2}{N}, t=T_{pred} \ 
% \end{equation}

\subsection{Implementation details}

We trained the model using a batch size of 256 for 100 epochs, employing the Stochastic Gradient Descent (SGD) optimizer to update the network parameters. To optimize training dynamics, we adopted the One-Cycle learning rate scheduler. The learning rate was initialized at 0.05 and gradually increased to a peak value of 0.2 before being reduced again, with updates applied at every batch iteration to ensure smooth convergence. The hidden dimension \( C \) was set to 32, and the traj-UGnet architecture consisted of 4 layers for both the downsampling and upsampling steps. 

For the diffusion model, we set the diffusion step count to 100 and fixed \( \beta_K \) at 0.05. A linear schedule was used for the variance schedule in the diffusion process, progressively adding noise to the data during training to guide the model toward better generalization. This configuration was designed to effectively capture the spatio-temporal dynamics of the data while maintaining training stability and efficiency.

\begin{table}[htb]
    \centering
    \caption{Evaluation metrics over 5,10,15 predicted time steps on real world AIS datasets: Tianjin Port(TJP) and Caofeidian Waters(CFW).}
    \label{table:comparison}
    \resizebox{\columnwidth}{!}{
    \begin{tabular}{>{\centering\arraybackslash}p{1.5cm}>{\centering\arraybackslash}p{1.5cm}>{\centering\arraybackslash}p{1.5cm}>{\centering\arraybackslash}p{1.5cm}>{\centering\arraybackslash}p{1.5cm}>{\centering\arraybackslash}p{1.5cm}>{\centering\arraybackslash}p{1.5cm}>{\centering\arraybackslash}p{1.5cm}>{\centering\arraybackslash}p{1.5cm}}
        \toprule
        Dataset & Step & Evaluation Metrics & LSTM\cite{tang2022model} & Seq2Seq\cite{forti2020prediction}  & Social-stgcnn\cite{mohamed2020social} & Social-GAN\cite{gupta2018social} & MID\cite{gu2022stochastic} & \textbf{STGDPM} \\
        \midrule
        \multirow{6}{*}{TJP} & \multirow{2}{*}{5} & ADE & 0.598 & 0.566 & 0.080 & 0.073 & 0.057 & \textbf{0.046} \\
        & & FDE & 0.879 & 0.805 & 0.090 & 0.119 & 0.080 & \textbf{0.060} \\
        \cmidrule{2-9}
        & \multirow{2}{*}{10} & ADE & 1.278 & 1.232 & 0.146 & 0.138 & 0.101 & \textbf{0.089} \\
        & & FDE & 2.314 & 2.262 & 0.218 & 0.259 & 0.183 & \textbf{0.136} \\
        \cmidrule{2-9}
        & \multirow{2}{*}{15} & ADE & 2.063 & 2.040 & 0.224 & 0.214 & 0.158 & \textbf{0.142} \\
        & & FDE & 3.981 & 4.052 & 0.386 & 0.439 & 0.305 & \textbf{0.240} \\
        \midrule
        \multirow{6}{*}{CFW} & \multirow{2}{*}{5} & ADE & 0.475 & 0.479 & 0.105 & 0.073 & 0.063 & \textbf{0.054} \\
        & & FDE & 0.693 & 0.678 & 0.113 & 0.102 & 0.083 & \textbf{0.069} \\
        \cmidrule{2-9}
        & \multirow{2}{*}{10} & ADE & 0.933 & 0.937 & 0.175 & 0.113 & 0.097 & \textbf{0.092} \\
        & & FDE & 1.557 & 1.544 & 0.233 & 0.185 & 0.162 & \textbf{0.134} \\
        \cmidrule{2-9}
        & \multirow{2}{*}{15} & ADE & 1.400 & 1.407 & 0.247 & 0.158 & 0.142 & \textbf{0.130} \\
        & & FDE & 2.456 & 2.448 & 0.368 & 0.288 & 0.240 & \textbf{0.205} \\
        \midrule
        \multirow{2}{*}{AVG} & \multirow{2}{*}{} & ADE & 1.125 & 1.110 & 0.163 & 0.128 & 0.103 & \textbf{0.094} \\
        & & FDE & 1.980 & 1.965 & 0.235 & 0.232 & 0.176 & \textbf{0.143} \\
        \bottomrule
    \end{tabular}}
\end{table}

\subsection{Quantitative evaluation}

Two evaluation metrics, ADE and FDE, are used to quantitatively assess the trajectory prediction results, which are compared in Table~\ref{table:comparison}, the units of evaluation metrics are standardized to 100 meters for consistent comparison across methods. These metrics reflect the performance of each method. LSTM and Seq2Seq models struggle to effectively learn trajectory features, leading to the lowest quality predictions. Social-STGCNN, which leverages a Graph Neural Network (GNN) to extract trajectory features, shows improved prediction accuracy and robustness compared to simpler networks. Social-GAN, which incorporates a Generative Adversarial Network (GAN), also achieves better prediction quality. MID, using a combination of Diffusion Models (DDPM) and Transformers, delivers promising results. In contrast, STGDPM outperforms all other methods, providing the most accurate and robust predictions under all experimental conditions. By learning the dynamic interactions between neighboring vessels and leveraging the diffusion model's ability to generalize, STGDPM significantly enhances both the accuracy and robustness of trajectory prediction.

\begin{figure}
\includegraphics[width=\textwidth]{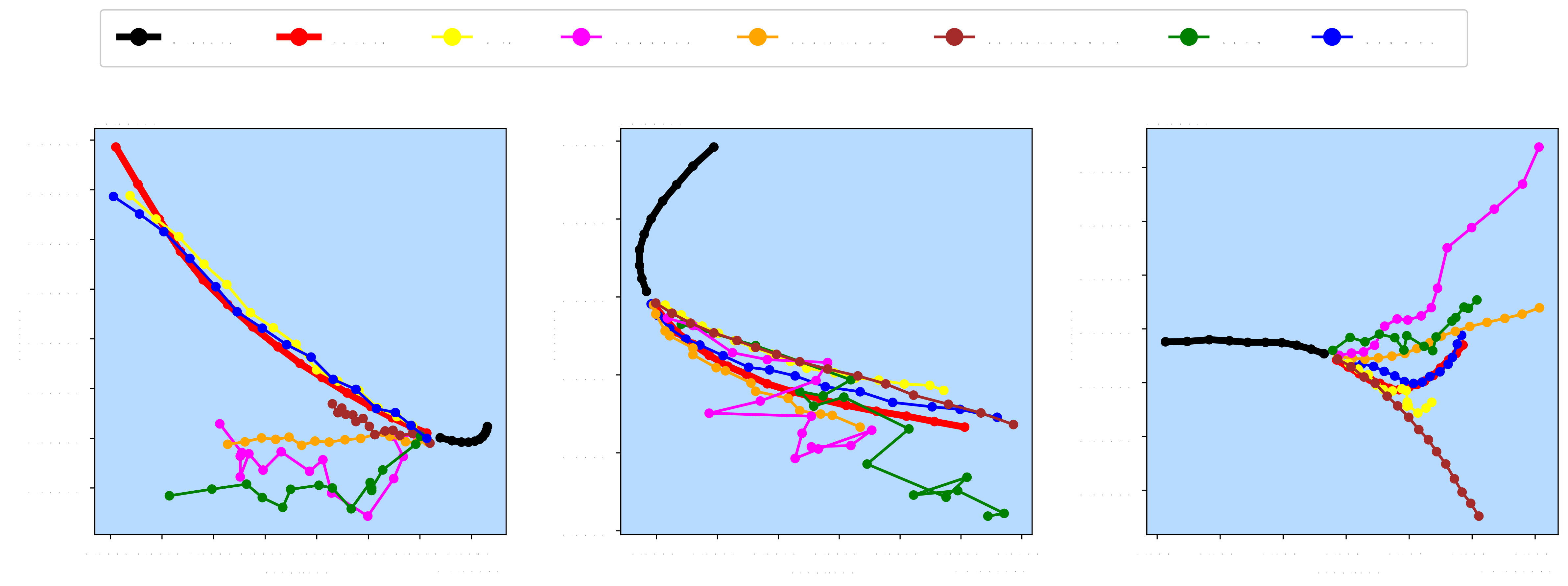}
\caption{We compared the prediction results of various methods in complex scenarios. The left image shows a turning and acceleration scenario, while the middle and right images depict turning scenarios.} 
\label{fig:comparison}
\end{figure}

\subsection{Qualitative evaluation}

To further evaluate the performance of the prediction method based on the large AIS dataset, vessel trajectories exhibiting complex behavior characteristics were selected for experimental analysis. Fig~\ref{fig:comparison} show a visual comparison of different competing methods. As shown, the predicted trajectories generated by LSTM and Seq2Seq deviate significantly from the actual trajectories. These models struggle to capture complex movement patterns, resulting in poor alignment with the true future trajectory. Social-STGCNN and Social-GAN exhibits better alignment with the future trajectory compared to simpler models. However, it still falls short in predicting subtle trajectory variations, Social-STGCNN and Social-GAN produce relatively lower-quality predictions, particularly when speed and heading change drastically. For instance, in the leftmost scene of Fig. 4, these models fail to capture the vessel’s acceleration behavior. In contrast, the MID model demonstrates promising performance, with predictions that closely follow the actual trajectory. The STGDPM, leveraging the STG feature, excels in predicting vessel trajectories with high accuracy and robustness, consistently provides the most accurate and reliable trajectory predictions. Across all subplots, STGDPM demonstrates the lowest deviation from the ground truth.

\begin{figure}
\includegraphics[width=\textwidth]{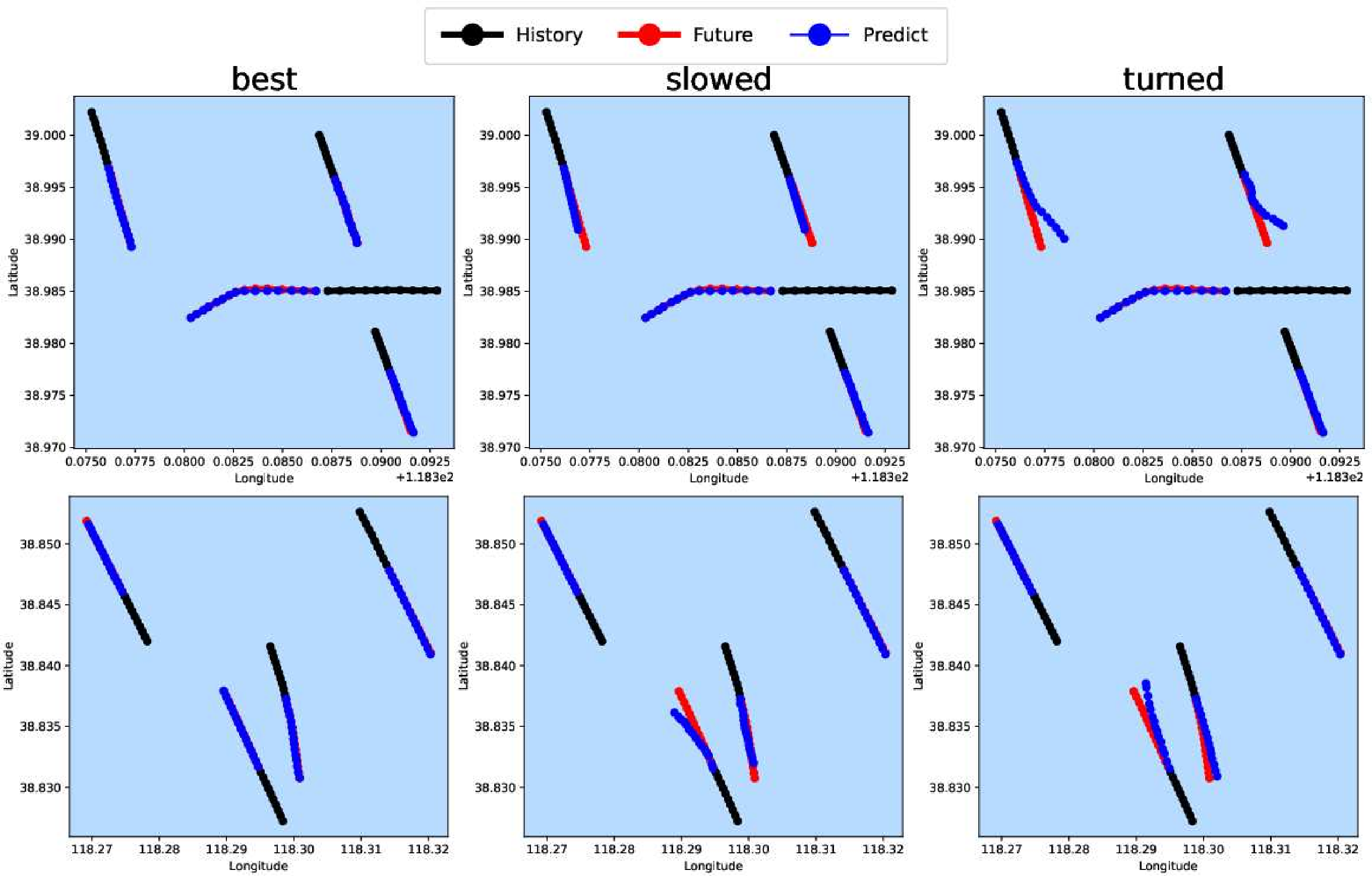}
\caption{Behavioral multi-modality of trajectories predicted by our model.} 
\label{fig:multimodal}
\end{figure}

Fig~\ref{fig:multimodal}, demonstrates the multimodal nature of the behaviors learned by the model in multi-ship interactive scenarios. From left to right, the predictions are categorized as best, showed, and turned, where the leftmost is the prediction most aligned with the ground truth, the middle shows a deceleration and yielding behavior, and the rightmost depicts a turning behavior. In the first row of Fig~\ref{fig:multimodal}, the ship traveling horizontally follows the "starboard rule," slowing down and changing course to give way. According to the International Regulations for Preventing Collisions at Sea (COLREGs), even a vessel with the right of way in a crossing situation must take avoiding action when necessary to prevent a collision. The top-right scene in Figure 3 illustrates the turning prediction based on this collision avoidance rule, while the bottom-right scene shows the turning prediction following the "starboard rule."

The results show that for trajectory prediction with complex behavioral characteristics, the STG method combined with the diffusion model is effective. Therefore, predicting the trajectory through STGDPM can improve the safety and efficiency of maritime traffic in complex navigation environments.

\subsection{Ablation Studies}
\label{Ablation}

\begin{minipage}{0.52\textwidth}
\justifying

\subsubsection{Effectiveness of Key Modules.} We perform an ablation study by removing key components from our final model: Unet, DynamicGraphConv (DGC), and the residual operation (Res) in the residual block, as shown in Table~\ref{table:ablation}. The addition of the Unet structure improved the results by 64.4\% for ADE and 51.7\% for FDE. With the inclusion of DynamicGraphConv, the improvements increased to 66.2\% for ADE and 57.1\% for FDE. 

\end{minipage}\hfill
\begin{minipage}{0.45\textwidth}
    \centering
    \captionsetup{justification=raggedright, width=\linewidth}
    \captionof{table}{Ablation studies: ADE, FDE results are the average over two tested scenes.}
    \label{table:ablation}
    \begin{tabular}{cccccc}
        \toprule
        Unet & DGC &  Res & ADE$\downarrow$  & FDE$\downarrow$    \\
        \midrule
        & & &  0.315 & 0.385   \\
        \checkmark & &  & 0.113 & 0.186   \\
        \checkmark &\checkmark & & 0.107 & 0.165  \\
        \checkmark & \checkmark  & \checkmark & 0.094 & 0.143  \\
        \bottomrule
    \end{tabular}
\end{minipage}
Finally, when all components are included, the results show improvements of 70.1\% for ADE and 62.8\% for FDE, respectively. 

\noindent\begin{minipage}[t]{0.48\textwidth} % 左栏
\vspace{0pt} % 强制顶部对齐
\justifying % 两端对齐
\subsubsection{Effectiveness of Interaction Boundary.} We conducted extensive experiments on the interaction boundary $\tau$ of the equation~\ref{eq:kernel} to investigate how different values of $\tau$  affect prediction accuracy. The experimental results are shown in Fig~\ref{fig:tau}, where the ADE and FDE values represent the average results across two datasets. 
\end{minipage}\hfill
\begin{minipage}[t]{0.48\textwidth} % 右栏
\vspace{0pt} % 强制顶部对齐
\centering
\includegraphics[width=\linewidth]{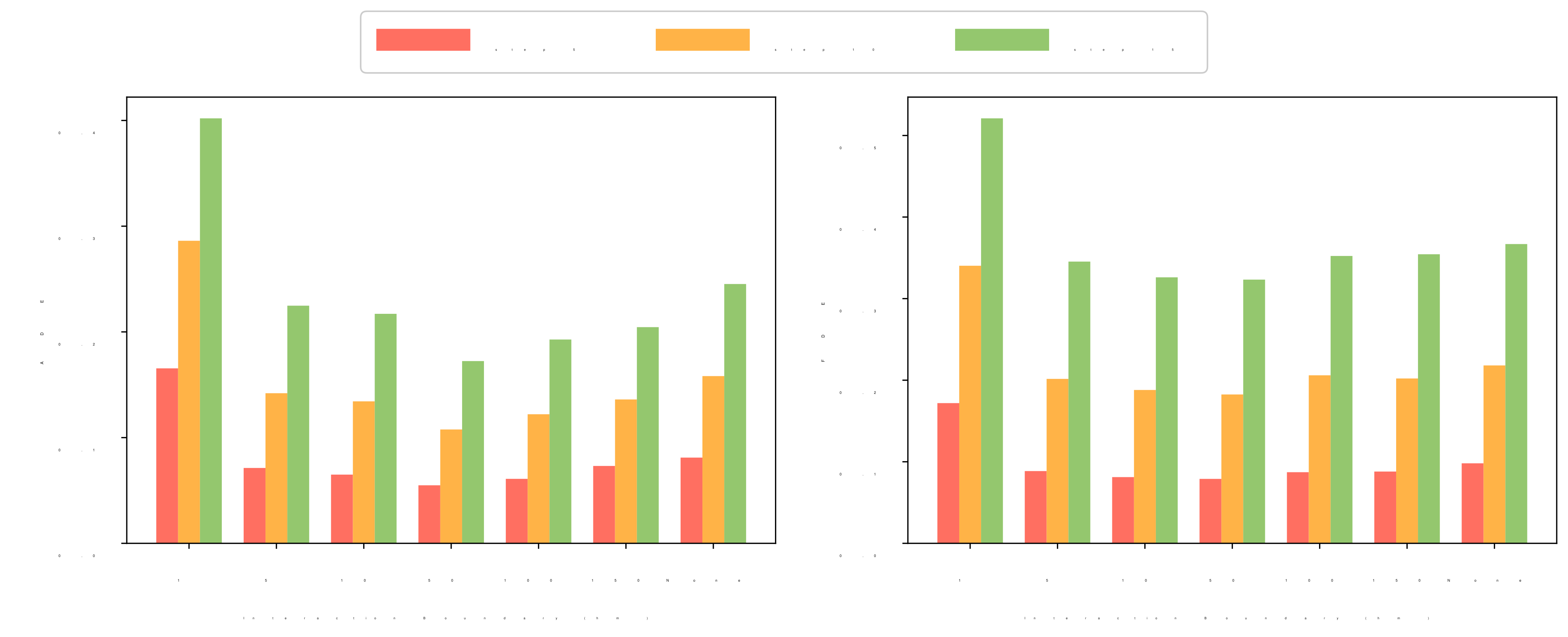}
\captionof{figure}{The impact of interaction boundary on ADE/FDE.}
\label{fig:tau}
\end{minipage}
The interaction boundary is measured in hectometers (hm). As $\tau$  increases from 1, both ADE and FDE initially decrease and then increase, with the optimal results achieved at $\tau$  = 50 . This suggests that too small a value of $\tau$  prevents the model from extracting spatial information from interactive scenes, while too large a value of $\tau$  may hinder the model’s judgment. As $\tau$  increases, the performance gradually approaches that of the case where no interaction boundary is set (represented as "None" in the Fig~\ref{fig:tau}).

\section{Conclusion}

In this paper, a Spatio-Temporal Graphs Diffusion Probabilistic Model framework is proposed for the challenges of trajectory prediction in the context of spatio-temporal data. Our approach integrates graph neural networks with advanced diffusion processes to capture complex dynamic spatio-temporal dependencies, effectively modeling the diffusion of trajectories over time. The proposed method holds great potential for maritime applications, including autonomous navigation, intelligent collision avoidance, and abnormal behavior detection. These capabilities highlight its practical value in enhancing safety and efficiency in maritime traffic management. In future work, we aim to explore further enhancements in model scalability, interpretability, and the integration of diffusion-based STGNNs with emerging technologies to expand their applicability in real-world trajectory prediction tasks.

\bibliographystyle{splncs04}
\bibliography{mybibliography}

\begin{thebibliography}{10}
\providecommand{\url}[1]{\texttt{#1}}
\providecommand{\urlprefix}{URL }
\providecommand{\doi}[1]{https://doi.org/#1}

\bibitem{akyuz2015hybrid}
Akyuz, E.: A hybrid accident analysis method to assess potential navigational contingencies: The case of ship grounding. Safety science  \textbf{79},  268--276 (2015)

\bibitem{alahi2016social}
Alahi, A., Goel, K., Ramanathan, V., Robicquet, A., Fei-Fei, L., Savarese, S.: Social lstm: Human trajectory prediction in crowded spaces. In: Proceedings of the IEEE conference on computer vision and pattern recognition. pp. 961--971 (2016)

\bibitem{capobianco2021deep}
Capobianco, S., Millefiori, L.M., Forti, N., Braca, P., Willett, P.: Deep learning methods for vessel trajectory prediction based on recurrent neural networks. IEEE Transactions on Aerospace and Electronic Systems  \textbf{57}(6),  4329--4346 (2021)

\bibitem{dendorfer2021mg}
Dendorfer, P., Elflein, S., Leal-Taix{\'e}, L.: Mg-gan: A multi-generator model preventing out-of-distribution samples in pedestrian trajectory prediction. In: Proceedings of the IEEE/CVF International Conference on Computer Vision. pp. 13158--13167 (2021)

\bibitem{feng2022stgcnn}
Feng, H., Cao, G., Xu, H., Ge, S.S.: Is-stgcnn: An improved social spatial-temporal graph convolutional neural network for ship trajectory prediction. Ocean Engineering  \textbf{266},  112960 (2022)

\bibitem{forti2020prediction}
Forti, N., Millefiori, L.M., Braca, P., Willett, P.: Prediction oof vessel trajectories from ais data via sequence-to-sequence recurrent neural networks. In: ICASSP 2020-2020 IEEE International Conference on Acoustics, Speech and Signal Processing (ICASSP). pp. 8936--8940. IEEE (2020)

\bibitem{gu2022stochastic}
Gu, T., Chen, G., Li, J., Lin, C., Rao, Y., Zhou, J., Lu, J.: Stochastic trajectory prediction via motion indeterminacy diffusion. In: Proceedings of the IEEE/CVF Conference on Computer Vision and Pattern Recognition. pp. 17113--17122 (2022)

\bibitem{gupta2018social}
Gupta, A., Johnson, J., Fei-Fei, L., Savarese, S., Alahi, A.: Social gan: Socially acceptable trajectories with generative adversarial networks. In: Proceedings of the IEEE conference on computer vision and pattern recognition. pp. 2255--2264 (2018)

\bibitem{han2019real}
Han, P., Wang, W., Shi, Q., Yang, J.: Real-time short-term trajectory prediction based on gru neural network. In: 2019 IEEE/AIAA 38th Digital Avionics Systems Conference (DASC). pp.~1--8. IEEE (2019)

\bibitem{ho2020denoising}
Ho, J., Jain, A., Abbeel, P.: Denoising diffusion probabilistic models. Advances in neural information processing systems  \textbf{33},  6840--6851 (2020)

\bibitem{kaluza2010complex}
Kaluza, P., K{\"o}lzsch, A., Gastner, M.T., Blasius, B.: The complex network of global cargo ship movements. Journal of the Royal Society Interface  \textbf{7}(48),  1093--1103 (2010)

\bibitem{khosroshahi2016surround}
Khosroshahi, A., Ohn-Bar, E., Trivedi, M.M.: Surround vehicles trajectory analysis with recurrent neural networks. In: 2016 IEEE 19th International Conference on Intelligent Transportation Systems (ITSC). pp. 2267--2272. IEEE (2016)

\bibitem{liu2023qsd}
Liu, R.W., Hu, K., Liang, M., Li, Y., Liu, X., Yang, D.: Qsd-lstm: Vessel trajectory prediction using long short-term memory with quaternion ship domain. Applied Ocean Research  \textbf{136},  103592 (2023)

\bibitem{liu2022deep}
Liu, R.W., Liang, M., Nie, J., Lim, W.Y.B., Zhang, Y., Guizani, M.: Deep learning-powered vessel trajectory prediction for improving smart traffic services in maritime internet of things. IEEE Transactions on Network Science and Engineering  \textbf{9}(5),  3080--3094 (2022)

\bibitem{liu2022stmgcn}
Liu, R.W., Liang, M., Nie, J., Yuan, Y., Xiong, Z., Yu, H., Guizani, N.: Stmgcn: Mobile edge computing-empowered vessel trajectory prediction using spatio-temporal multigraph convolutional network. IEEE Transactions on Industrial Informatics  \textbf{18}(11),  7977--7987 (2022)

\bibitem{ma2024dm}
Ma, K., Han, Q., Yao, J., Zhang, Y.: Dm-bdd: Real-time vessel trajectory prediction based on diffusion probability model balancing diversity and determinacy. In: 2024 International Joint Conference on Neural Networks (IJCNN). pp.~1--8. IEEE (2024)

\bibitem{mohamed2020social}
Mohamed, A., Qian, K., Elhoseiny, M., Claudel, C.: Social-stgcnn: A social spatio-temporal graph convolutional neural network for human trajectory prediction. In: Proceedings of the IEEE/CVF conference on computer vision and pattern recognition. pp. 14424--14432 (2020)

\bibitem{sadeghian2019sophie}
Sadeghian, A., Kosaraju, V., Sadeghian, A., Hirose, N., Rezatofighi, H., Savarese, S.: Sophie: An attentive gan for predicting paths compliant to social and physical constraints. In: Proceedings of the IEEE/CVF conference on computer vision and pattern recognition. pp. 1349--1358 (2019)

\bibitem{salzmann2020trajectron++}
Salzmann, T., Ivanovic, B., Chakravarty, P., Pavone, M.: Trajectron++: Dynamically-feasible trajectory forecasting with heterogeneous data. In: Computer Vision--ECCV 2020: 16th European Conference, Glasgow, UK, August 23--28, 2020, Proceedings, Part XVIII 16. pp. 683--700. Springer (2020)

\bibitem{simeunovic2021spatio}
Simeunovi{\'c}, J., Schubnel, B., Alet, P.J., Carrillo, R.E.: Spatio-temporal graph neural networks for multi-site pv power forecasting. IEEE Transactions on Sustainable Energy  \textbf{13}(2),  1210--1220 (2021)

\bibitem{sohl2015deep}
Sohl-Dickstein, J., Weiss, E., Maheswaranathan, N., Ganguli, S.: Deep unsupervised learning using nonequilibrium thermodynamics. In: International conference on machine learning. pp. 2256--2265. PMLR (2015)

\bibitem{tang2022model}
Tang, H., Yin, Y., Shen, H.: A model for vessel trajectory prediction based on long short-term memory neural network. Journal of Marine Engineering \& Technology  \textbf{21}(3),  136--145 (2022)

\bibitem{tekin2021spatio}
Tekin, S.F., Karaahmetoglu, O., Ilhan, F., Balaban, I., Kozat, S.S.: Spatio-temporal weather forecasting and attention mechanism on convolutional lstms. arXiv preprint arXiv:2102.00696  \textbf{4} (2021)

\bibitem{wen2023diffstg}
Wen, H., Lin, Y., Xia, Y., Wan, H., Wen, Q., Zimmermann, R., Liang, Y.: Diffstg: Probabilistic spatio-temporal graph forecasting with denoising diffusion models. In: Proceedings of the 31st ACM International Conference on Advances in Geographic Information Systems. pp. 1--12 (2023)

\bibitem{wu2024gl}
Wu, Y., Yv, W., Zeng, G., Shang, Y., Liao, W.: Gl-stgcnn: Enhancing multi-ship trajectory prediction with mpc correction. Journal of Marine Science and Engineering  \textbf{12}(6), ~882 (2024)

\bibitem{wu2020comprehensive}
Wu, Z., Pan, S., Chen, F., Long, G., Zhang, C., Philip, S.Y.: A comprehensive survey on graph neural networks. IEEE transactions on neural networks and learning systems  \textbf{32}(1),  4--24 (2020)

\bibitem{wu2019graph}
Wu, Z., Pan, S., Long, G., Jiang, J., Zhang, C.: Graph wavenet for deep spatial-temporal graph modeling. arXiv preprint arXiv:1906.00121  (2019)

\bibitem{yu2017spatio}
Yu, B., Yin, H., Zhu, Z.: Spatio-temporal graph convolutional networks: A deep learning framework for traffic forecasting. arXiv preprint arXiv:1709.04875  (2017)

\bibitem{yuan2021agentformer}
Yuan, Y., Weng, X., Ou, Y., Kitani, K.M.: Agentformer: Agent-aware transformers for socio-temporal multi-agent forecasting. In: Proceedings of the IEEE/CVF International Conference on Computer Vision. pp. 9813--9823 (2021)

\end{thebibliography}

\end{document}